# Decade-long Emission Forecasting with an Ensemble Model in Taiwan

Gordon Hung [1,*], Salinna Abdullah [2]

1. Hsinchu County American School, Independent Researcher; gordonh741@gmail.com
2. University College London, Independent Researcher; salinna.abdullah.13@ucl.ac.uk
* Correspondence: gordonh741@gmail.com

**Abstract**

Taiwan's high population and heavy dependence on fossil fuels have led to severe air pollution, with the most prevalent greenhouse gas being carbon dioxide ($CO_2$). Therefore, this study presents a reproducible and comprehensive case study comparing 21 of the most commonly employed time series models in forecasting emissions, analyzing both univariate and multivariate approaches. Among these, Feedforward Neural Network (FFNN), Support Vector Machine (SVM), and Random Forest Regressor (RFR) achieved the best performances. To further enhance robustness, the top performers were integrated with Linear Regression through a custom stacked generalization ensemble technique. Our proposed ensemble model achieved an SMAPE of 1.407 with no signs of overfitting. Finally, this research provides an accurate decade-long emission projection that will assist policymakers in making more data-driven decisions.

## 1. Introduction

Climate change is a pressing concern in the contemporary world [1, 2]. With extreme changes in weather conditions leading to unpredictable natural disasters like flash floods and hazardous smog events [3, 4]. Oceans and lakes are also becoming more acidic, causing many marine life to lose their habitats [5, 6]. These environmental issues are closely tied to the escalating levels of $CO_2$ emissions, which are one of the biggest contributors to global warming and climate disruptions [7, 8].

Taiwan is one of the top emitters of $CO_2$ in the world due to its rapidly developing economy [9, 10]. Taiwan's emissions are largely contributed by its energy sector, which includes multiple renewable and nonrenewable sources, accounting for over 70 percent of total annual emissions [11]. Among the energy sources, fossil fuels like coal, oil, and natural gas are the greatest contributing factors [12]. According to the Ministry of Environment, Taiwan emitted over 257 million metric tons of $CO_2$ in 2021, showing an increase in emissions over the past decade [13]. However, it is challenging for Taiwan to significantly reduce its $CO_2$ emissions in a short period because its economy is heavily reliant on manufacturing and production [14]. Furthermore, most of Taiwan's major industries rely on fossil fuels, which have led to a drastic increase in $CO_2$ emissions.

Additionally, Taiwan's high population density has resulted in higher energy demands, leading to greater emissions from residential and commercial buildings [15, 16]. The continuous increase in emission levels in Taiwan has increased the health risks of its population, leading to more cases of respiratory diseases [17]. Environmental degradation is also a detrimental effect of $CO_2$ emissions. For example, water and soil quality deteriorate after prolonged exposure to high levels of $CO_2$ [18, 19]. Land and marine animals also suffer from the loss of habitats and food sources, reducing the biodiversity in Taiwan. These problems can also affect the economy, particularly sectors that are sensitive to changes in weather patterns, like agriculture and fisheries [20, 21].

Taiwan's government has implemented various policies to reduce CO2 emissions and mitigate its effects. For example, the government updated the Nationally Determined Contribution (NDC) under the Paris Agreement in 2021 to set new goals for the upcoming years [22]. The updated NDC aims for a reduction of 50 percent by 2030 compared to 2005 levels. Enacted in 2009 and amended in 2019, the Renewable Energy Development Act promotes the reliance on renewable energy sources by incentivizing solar and wind energy developments [23]. In 2022, Taiwan officially established the goal of achieving net-zero emissions in 2050 [24]. The government has been collaborating with local companies to encourage reliance on clean energy sources. As a result, the total CO2 emissions in Taiwan have been increasing at a much slower rate compared to previous years [25, 26]. However, while these policies are effective, more measures and initiatives must be taken in order to reach the aforementioned goals.

The remainder of this paper is organized as follows: Section 2 provides a comprehensive literature review of past works. Section 3 discusses data preprocessing, model selection, and the hybrid model architecture. Section 4 offers the future forecasts of the optimized hybrid ensemble model. Finally, this paper concludes in Section 5.

## 2. Related Work

Forecasting CO2 emissions is a crucial aspect of energy development policies in various countries. Different models have been developed to predict CO2 emissions, with some focusing on univariate time series data while others utilize multivariate models [27, 28].

The study by Alharbi et al. utilizes feed-forward neural networks (FFNN), adaptive network-based fuzzy inference systems (ANFIS), and long short-term memory (LSTM) to forecast CO2 emissions in Saudi Arabia, achieving a 93.13% accuracy. The ensemble prediction indicates a decrease in emissions from 9.4976 million tonnes in 2020 to 6.1707 million tonnes by 2030 [29]. Kumar et al. analyzed India's CO2 emissions, predicting the next decade's trends using univariate time-series data from 1980 to 2019 with models including ARIMA, SARIMAX, Holt-Winters, linear regression, random forest, and LSTM. The findings indicate that the LSTM model outperforms others with a 3.101% MAPE and 60.635 RMSE [30]. Khan et al. conducted a comparative analysis of linear and nonlinear time series models to forecast CO2 emissions in Pakistan; they found that nonlinear machine learning models yield the lowest RMSE and MAE values [31]. Sharma et al. analyzed CO2 emissions in India from 1995 to 2018 using multiple linear regression, with year, population, and electricity consumption as independent variables. The model achieved a test score of 96.40%, highlighting the significant impact of fossil fuel consumption on environmental issues [32]. Ali et al. proposed a hybrid forecasting technique for CO2 emissions in Pakistan by analyzing the long-run trend and residual subseries using parametric and nonparametric regression methods alongside standard time series models. The results indicate that this hybrid approach is highly accurate, achieving an MAPE score of 3.762 [33]. Zhang et al. evaluated the forecasting performance of autoregressive (AR), seasonal autoregressive integrated moving average (SARIMA), and threshold autoregressive (SETAR) models, along with their hybridization with artificial neural networks (ANN), using Canadian lynx data. The findings indicate that while hybrid models do not always enhance forecasting accuracy, the SETAR-ANN combination demonstrates significant predictive capability, especially for generating up to 10-step forecasts [34]. Yılmaz et al. proposed a hybrid method for estimating CO2 emissions in Türkiye, combining the Shuffled Frog-Leaping Algorithm (SFLA) and the Firefly Algorithm (FA) to enhance optimization. This approach is integrated with an Artificial Neural Network (ANN), achieving an RMSE of 5.1107 and an R² of 0.9904 [35].

Overall, the literature suggests that hybrid time series models leverage advanced techniques such as neural networks and deep learning to improve prediction accuracy and inform energy development policies. These hybrid time series models often outperform regular models and are best suited for long-term time-series forecasting. Additionally, the integration of multiple variables in the forecasting process has shown promising results in predicting CO2 emissions in various regions. Table 1 shows a summary of related studies.

**Table 1.** Literature review.

| Ref | Best Model | Location | Accuracy |
|---|---|---|---|
| [29] | Ensemble Model: FFNN, ANFIS, LSTM | Saudi Arabia | MAPE: 6.8675% |
| [30] | LSTM | India | MAPE: 3.101% RMSE: 60.635 |
| [31] | NNAR(1,1) | Pakistan | RMSE: 0.0007 MAE: 0.0005 |
| [32] | Multiple Linear Regression | India | Test Score: 96.40% |
| [33] | Hybrid Forecasting Technique | Pakistan | MAPE: 3.762% |
| [34] | SETAR-ANN | Canada | MAE: 0.357 MAPE: 5.13% |
| [35] | Hybrid Method | Türkiye | RMSE: 5.11107 R2: 0.9904 |
| **NA** | **Ensemble Model: FFNN, SVR, RFR** | **Taiwan** | **MAE: 0.166 MAPE: 1.398%** |

While the current literature encompasses various powerful techniques, there are still clear research gaps that this paper seeks to address. First, none of the previous studies have conducted comprehensive evaluations on commonly used univariate and multivariate models. Second, while various studies have experimented with hybrid models, there has been limited research into integrating three black-box models. Third, there hasn't been a reproducible 10-year forecast in Taiwan, a heavily populated region that is representative of many Asian countries.

## 3. Methodology

Building upon previous studies, this paper utilizes a state-of-the-art approach to accurate emission forecasting. We employed 10 univariate and 11 multivariate time series models, with evaluation metrics such as MAE, MSE, RMSE, MAPE, Symmetric MAPE, and Max Error. Then, we utilized a custom stacked generalization technique to integrate our top performers with a meta-model. Lastly, the ensemble model was used to forecast CO2 emissions for the next 10 years: 2023 - 2032. The contributions of this paper are as follows:
- Provide a comprehensive study on univariate and multivariate time series models.
- Present a novel ensemble model that outperforms previously proposed models.
- Forecast CO2 emissions in Taiwan for the next 10 years.

The outline of the steps taken is shown in Figure 1.

**Figure 1.** This is a figure. Schemes follow the same formatting.

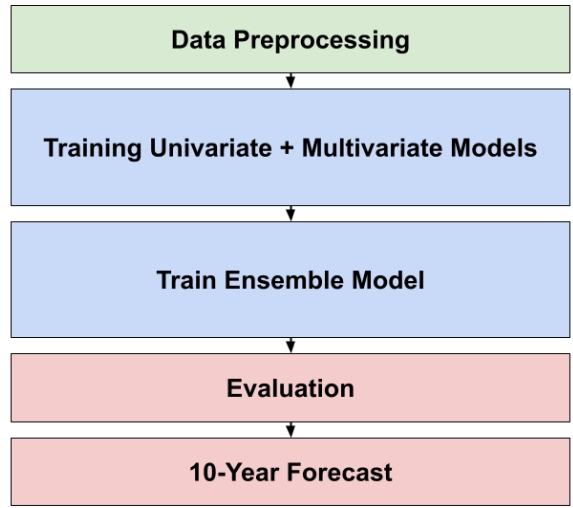

*3.1. Data Acquisition and Preprocessing*

The dataset for this study is a custom dataset acquired from Our World in Data [36]. Ranging from 1965 to 2022, it consists of annual data samples of CO2 emissions per capita in tons (t), total gas consumption, total coal consumption, and total oil consumption in terawatt-hours (TWh). The features are selected because the energy sector in Taiwan accounts for over 70 percent of direct CO2 emissions; the selected sources are also the biggest contributors within the energy sector. The dataset does not contain any NA values, and all samples are in numerical form. In this study, a lag value of 3 years was used for time series forecasting of CO2 emissions per capita. This lag value ensures the model captures the most relevant dependencies while not being overly influenced by unnecessary noise. Figure 2 provides a graphical representation of the features and target.

**Figure 2.** Dataset before preprocessing.

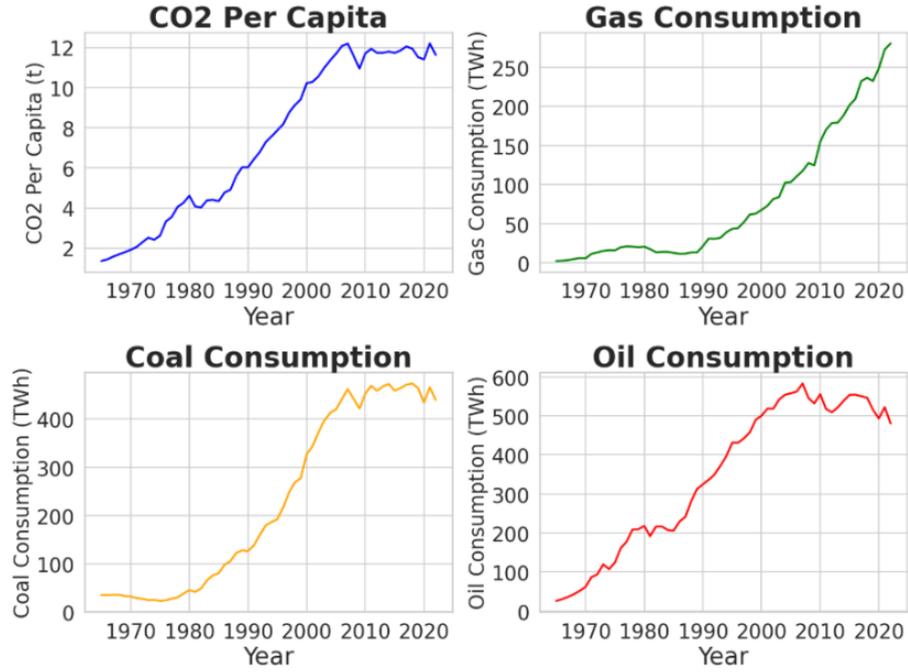

### 3.2. Stationarity

To determine the stationarity of our dataset, the Augmented Dickey-Fuller (ADF) Test and the Kwiatkowski-Phillips-Schmidt-Shin (KPSS) Test were applied to each of the columns. The ADF test is a commonly used statistical approach to check the presence of a unit root in a time series, which indicates non-stationarity. It builds on the basic Dickey-Fuller test by adding lagged differences of the time series to handle autocorrelation in the residuals. The ADF test can be described using Equation (1).

$$\Delta Y_t = \alpha + \beta t + \gamma Y_{t-1} + \sum_{i=1}^{p} \delta_i \Delta Y_{t-1} + \epsilon_t \qquad (1)$$

$\Delta Y_t$ is the first difference of the series $Y_t$. $\alpha$ is a constant. $\beta t$ is a time trend. $\gamma Y_{t-1}$ represents the coefficient on the lagged level of the series, which is very important in testing for a unit root. $\sum_{i=1}^{p} \delta_i \Delta Y_{t-1}$ represents the lagged differences to capture autocorrelation in the series. Lastly, $\epsilon_t$ represents white noise. The null hypothesis for the ADF test is that there is a unit root: $\gamma = 0$; on the other hand, an alternative hypothesis is that the series is stationary. The KPSS test is another statistical test that can be implemented to test the stationarity of the series. It is based on the residuals from the ordinary least squares (OLS) regression of the time series on an intercept or on both an intercept and time trend. The KPSS test can be described using Equation (2).

$$\text{KPSS} = \frac{1}{r^2} \sum_{t=1}^{T} \frac{S_t^2}{\widehat{\sigma^2}} \qquad (2)$$

T represents the number of observations, and $S_t$ represents the partial sum of the residuals at time t. $\widehat{\sigma^2}$ is the estimated long-run variance of the residuals. If the test statistic is greater than the critical value, the series may be non-stationary. According to the tests, none of the columns were stationary, so we applied the differencing algorithm for stationary conversion. We applied the differencing algorithm once to the columns CO2 and Oil, and twice to the columns Gas and Coal, as they remained non-stationary after the first application. After the stationary conversation, the first two values in the original series were lost because they didn't have a previous value to subtract from. Equation (3) represents the differencing algorithm.

$$\Delta Y_t = y_t - y_{t-1} \tag{3}$$

Where $\Delta Y_t$ is the differenced value at time t, $y_t$ is the value of the series at time t, and $y_{t-1}$ is the value at time t - 1. Figure 3 shows the features and the target after applying the differencing algorithm.

**Figure 3.** Dataset before preprocessing.

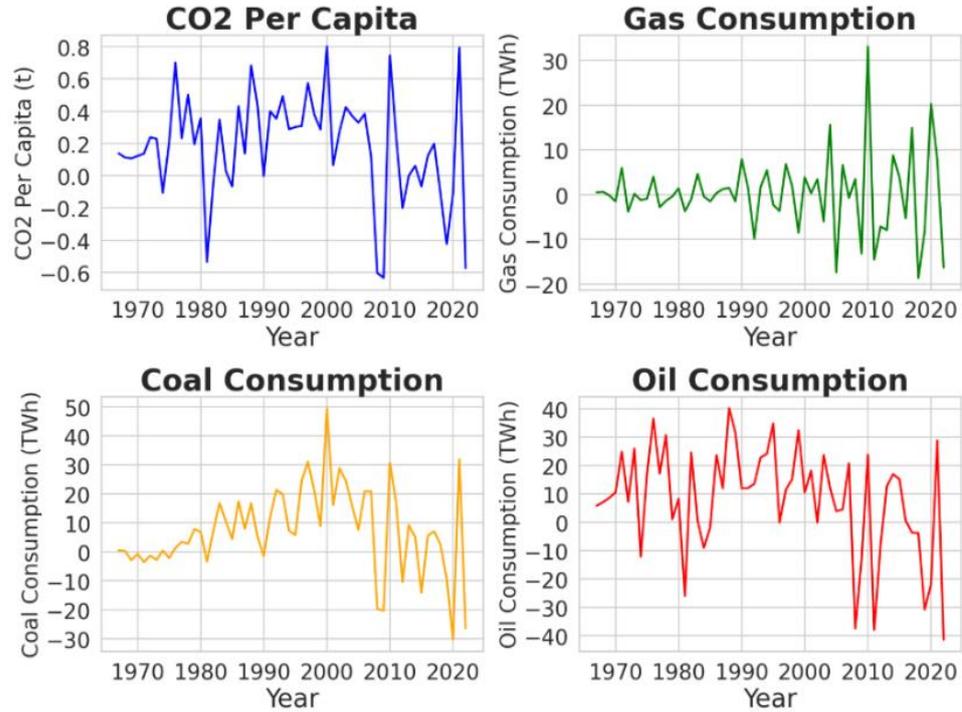

### 3.3. Multicollinearity

After making our dataset stationary, we checked for multicollinearity between the features. Multicollinearity is an undesirable situation in regression analysis where independent features are highly correlated with each other. This can result in poor performance and inaccurate evaluations. To detect multicollinearity, we used the Variance Inflation Factor (VIF) to measure the correlation between the features. A VIF value of 1 to 5 indicates little multicollinearity, whereas a VIF value of greater than 5 indicates high multicollinearity. Table 2 shows the VIF value of each feature.

**Table 2.** VIF value for each feature.

| Features | VIF Values |
|---|---|
| Gas | 1.171 |
| Coal | 1.566 |
| Oil | 1.391 |

As seen from Table 2, the features show little multicollinearity.

### 3.4. Data Split

After extensive preprocessing, we split our data into two sets: training and testing. The training set consists of 46 data samples from 1967 to 2012 inclusive, and the testing set consists of 10 data samples from 2013 to 2022 inclusive. This division of data samples is optimal because it aligns with the objective of our paper, which is to forecast 10 years into the future. Figure 4 shows the graphical representation of our training and testing sets.

**Figure 4.** Train and test split.

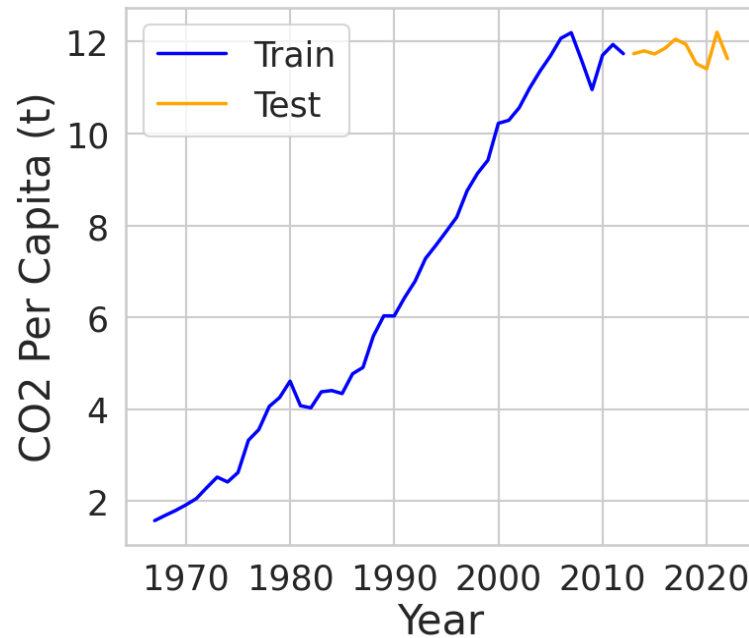

*3.5. Evaluation Metrics*

To accurately assess the performance of our models, we employed six evaluation matrices that provide insights into our models from different perspectives. We utilized Mean Absolute Error (MAE), Mean Squared Error (MSE), Root Mean Squared Error (RMSE), Mean Absolute Percent Error (MAPE), Symmetric Mean Absolute Percent Error (SMAPE), and Max Error to evaluate both the univariate and multivariate time series models.

*3.6. Univariate Models*

In the field of time series forecasting, univariate models are often employed as baseline models to compare against more complex multivariate models. They can be highly effective when capturing long-term trends and seasonality. In this paper, we implemented 10 univariate time series models to act as our baseline models: Autoregressive Integrated Moving Average (ARIMA), Holt-Winters Exponential Smoothing (Holt-Winters), Seasonal Autoregressive Integrated Moving Average (SARIMA), Auto-Regressive Integrated Moving Average (Auto ARIMA), Simple Exponential Smoothing (SES), Holt's Linear Trend Model (Holt's Linear), Theta, Fast Fourier Transform (FFT), and Naive Drift.

ARIMA models combine autoregressive (AR) terms, moving average (MA), and differencing to understand trends in time series. Hot-Winters extends exponential smoothing by incorporating both trend and seasonal components, making it suitable for forecasting seasonal patterns. SARIMA is a modification of ARIMA that incorporates both seasonal and non-seasonal components to forecast trends with periodic patterns. Auto ARIMA automatically selects the best ARIMA parameters for SARIMA using criteria such as AIC or BIC. Simple Exponential Smoothing is optimal for time series without much seasonality; it places more emphasis on recent data compared to past samples. Holt's Linear Trend Model is an extension of Simple Exponential Smoothing that can understand trends in the series. The Theta Model splits time series into components for trend and seasonality to more comprehensively understand the data. FFT decomposes the time series data into frequency components, making it optimal for understanding periodic patterns. The Naive Model is a simple model that assumes the next observation will be the same as the last. The Drift Model is an extension of the Naive Model; it assumes that changes in the series will continue at the same average rate as past values.

*3.7. Multivariate Models*

After defining the univariate time series models as our base models, we implemented 11 multivariate time series models: Dynamic Factor Model (DFM), Long Short-Term Memory (LSTM), Feedforward Neural Network (FFNN), Vector Autoregression (VAR), Bayesian Vector Autoregression (BVAR), Vector Error Correction Model (VECM), Ridge Regression, Elastic Net Regression, Support Vector Regression (SVR), Random Forest Regression (RFR), and Decision Tree. These models are more complex and can understand the relationship between CO2 and other external factors.

DFM reduces the complexities of the data by finding the common trends in a large set of variables. While this can be effective in preventing overfitting, it can be detrimental to the model's learning. LSTM is a type of Recurrent Neural Network (RNN) that can handle time series data by maintaining long-term dependencies. FFNN is a Neural Network (NN) in which the connections between nodes do not form a cycle. It is optimal for understanding complex patterns but requires stationary inputs. VAR models are effective at capturing linear interdependencies between time series variables; it's optimal when variables like CO2, coal, gas, and oil influence each other, allowing them to be forecasted simultaneously. BVAR is an extension of VAR models as it incorporates Bayesian techniques to improve the robustness of the forecasts. Next, VECM does not require a stationary time series; it is commonly used to analyze long-term equilibrium relationships and short-term changes. Ridge regression is an extension of linear regression that adds a regularization penalty to prevent overfitting by shrinking large coefficients. This approach is optimal for filtering out noisy data. Elastic Net Regression is a combination of Ridge and Lasso regression; it adds regularization terms to balance the benefits of both regressions. SVR is a modification of the Support Vector Machine that is used for regression tasks. It fits the best line within a margin of error, making it robust against outliers. RFR is an ensemble model that constructs multiple decision trees on different subsets of the data. This model improves the stability by averaging the results from all the trees created. Lastly, Decision Trees are simpler versions of RFR; they split data into subsets based on certain criteria to make forecasts. These models are effective at capturing non-linear relationships.

*3.8. Top Models*

After a comprehensive evaluation of all the univariate and multivariate time series models, we acquired the top three models to forecast the CO2 trend in Taiwan for the next 10 years. Based on the evaluations seen in the Results Section, the top three models are ranked as follows: (1) FFNN, (2) SVR, and (3) RFR. A detailed explanation and analysis of each model will be provided in this section.

3.8.1. Feedforward Neural Network (FFNN)

FFNN is an advanced machine-learning model that is part of the artificial neural networks (ANNs) family. It is characterized by the flow of information only in one direction—from the input layer, through one or more hidden layers, to the output layer. This unique architecture is unlike other recurrent neural networks (RNNs) that utilize cycles in their connections to generate predictions. FFNN consists of interconnected neurons arranged in different layers. The first layer is the input layer, where each neuron in this layer represents a specific feature in the input data. The last layer is the output layer, where the number of neurons corresponds to the number of outputs. In between the input layer and the output layer are the hidden layers. These layers are more complex and are responsible for processing and understanding the dataset. With forward passes, FFNN can capture patterns and trends in the dataset and produce accurate forecasts.

During each forward pass, neurons in the layer calculate the weighted sum of the inputs, apply the activation function, and pass the result to the next layer. The mathematical representation of a neuron calculating the weighted sum of the inputs in a hidden layer is shown in Equation (4).

$$u_k = \sum_{j=i}^{n} x_j w_{jk} + b_k \qquad (4)$$

Where $u_k$ represents the weighted sum of the inputs for neuron k, and $x_j$ represents the $j^{th}$ input to the neuron. $w_{jk}$ is the weight associated with the $j^{th}$ input of neuron k. $b_k$ is the bias term for the neuron. After calculating the weighted sum, the neuron passes it through an activation function to introduce non-linearity into the model. This process can be represented by Equation (5).

$$j_k = activation(u_k) \qquad (5)$$

Figure 5 shows a fully connected FFNN, illustrating the general structure and flow of information within the model.

**Figure 5.** Conceptual representation of FFNN.

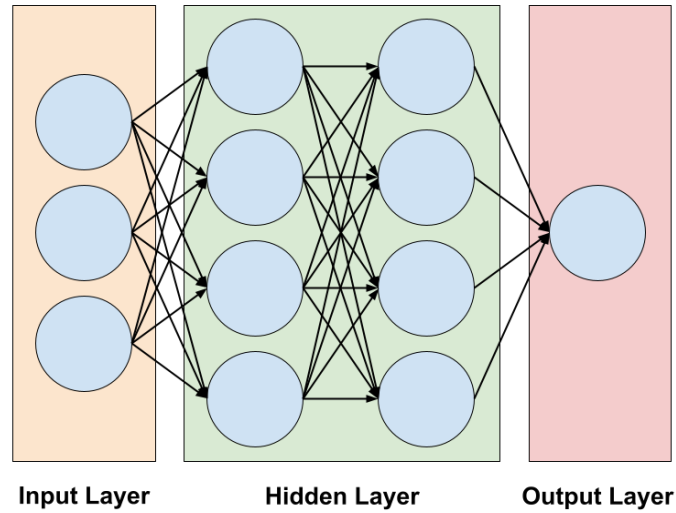

3.8.2. Support Vector Regressor (SVR)

SVR is an extension of Support Vector Machines (SVM) that is designed for regression machine learning problems. The theory of SVR is to construct a function that best fits the data while allowing some degree of error; this ensures a balance of accuracy and complexity. This is achieved by introducing a margin of tolerance, defined by the parameter $\epsilon$, which allows slight deviations from the actual values without causing a detrimental effect on the model. The loss function is defined in Equation (6).

$$L_{SVR}(\beta) = \sum_{i=1}^{n} \max_{0}(|y_i - f(x_i)| - \epsilon) + \lambda\|\beta\|^2 \qquad (6)$$

Where n represents the number of training samples. $y_i$ represents the actual target value for the $i^{th}$ observation. $f(x_i)$ is the predicted value based on historical data. $\lambda$ is the regularization parameter controlling the trade-off between model complexity and training error. $\|\beta\|^2$ represents the flatness of the function. To account for non-linear relationships between the features and the target, SVR employs kernel functions. First, the model maps the input features into a higher dimensional space using a kernel function $K(x_i, x_j)$. This is represented in Equation (7).

$$\phi(x) = K(x, x_i) \tag{7}$$

Then, in order to find the best-fit function, the model attempts to minimize the following function:

$$\frac{1}{2}\|\beta\|^2 + C\sum_{i=1}^{n} \xi_i \tag{8}$$

Where C is the regularization parameter that controls the trade-off between maximizing the margin and minimizing the error. $\xi_i$ are the slack variables that allow some data points to deviate from the actual values. After optimizing the parameters, the model can predict new data points as follows:

$$f(x) = \sum_{i=1}^{n}(a_i - a_i^*)K(x_i, x) + b \tag{9}$$

Where $a_i$ and $a_i^*$ are the Lagrange multipliers from the optimization process, representing the importance of the support vectors. b is the bias term. Figure 6 shows an architecture diagram of the SVR algorithm.

**Figure 6.** Conceptual representation of SVR.

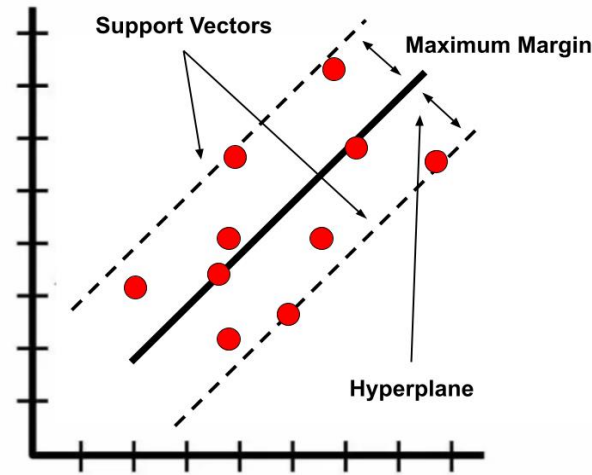

3.8.3. Random Forest Regressor (RFR)

RFR is an ensemble machine-learning model that utilizes predictions of multiple decision trees to generate a more accurate prediction. It prevents overfitting by averaging the predictions from individual decision trees, leveraging the diversity among the trees to improve the model's performance. The process of averaging the outputs of each decision tree is shown in Equation (10).

$$\hat{y} = \frac{1}{T}\sum_{t=1}^{T} f_t(x) \tag{10}$$

Where T represents the total number of trees in the forest. $f_t(x)$ represents the prediction from the $t^{th}$ tree. $\hat{y}$ represents the final ensemble prediction for input x. RFR is a relatively complex model that involves multiple steps. First, it utilizes bootstrap sampling to generate different training subsets. This ensures that each tree is trained on slightly different data, taking advantage of the variations in the forest. The mathematical representation of this process is shown below.

$$D_b \sim Bootstrap(D) \tag{11}$$

Here, $D_b$ represents a bootstrapped sample from the full dataset D. For each bootstrapped sample $D_b$, an individual decision tree $T_b$ is constructed. During the construction of each tree, only a random subset of features m is considered at each split, where m < p, with p representing the total number of features. This improves the model's ability to generalize by allowing greater variations within its structure. The mathematical representation of this process is shown below.

$$T_b = DecisionTree(D_b, m) \tag{12}$$

Where $T_b$ represents the $b^{th}$ decision tree, and $D_b$ is the bootstrapped sample used to train the tree. m represents the subset of features randomly selected at each split. After training all the individual trees, each tree provides its prediction, $T_b(x)$, for the input x. To obtain the final ensemble prediction, F(x), the predictions from all the trees are averaged; this reduces variance and improves the model's generalizability. The final model output equation is given below.

$$F(x) = \frac{1}{B} \sum_{b=1}^{B} T_b(x) \tag{13}$$

Where B represents the total number of decision trees, and $T_b(x)$ is the prediction from the $b^{th}$ tree. The ensemble result is represented by F(x). Figure 7 shows the architecture diagram of the RFR algorithm.

**Figure 7.** Conceptual representation of RFR.

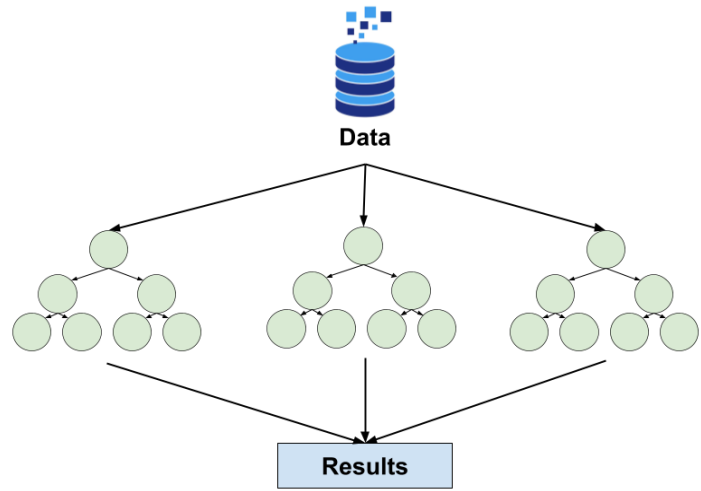

*3.9. Ensemble Model*
To optimize the top-performing models, we designed an ensemble model that utilizes stacked generalization to strategically combine the predictions of FFNN, SVR, and RFR. Stacked generalization is an ensemble learning technique that uses a meta-model to integrate the predictions of multiple base models. The ensemble model uses Linear Regression as the meta-model to calculate the optimal weight to assign to each model. A visualization of the ensemble model's architecture is shown in Figure 8.

**Figure 8.** Ensemble model architecture.

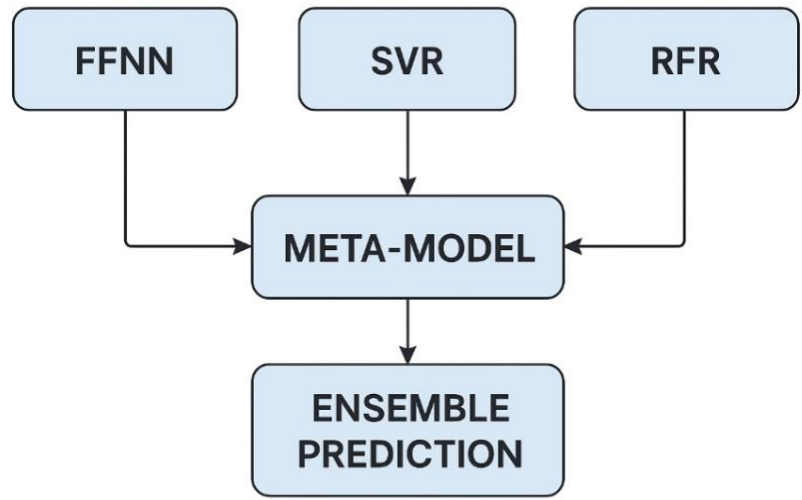

## 4. Results and Analysis

*4.1. Univariate Results*

After extensive hyperparameter optimizations, Table 3 presents the performances of the univariate models.

**Table 3.** Univariate model performances.

| Models | MAE | MSE | RMSE | MAPE | SMAPE | Max Error | Rank |
|---|---|---|---|---|---|---|---|
| **ARIMA** | **0.568** | **0.606** | **0.778** | **4.881** | **4.695** | **1.733** | **1** |
| **ARIMA** | **0.061** | **0.643** | **0.801** | **5.278** | **5.091** | **1.669** | **2** |
| **Simple Exp** | **0.716** | **0.786** | **0.887** | **6.117** | **2.921** | **1.589** | **3** |
| Holt Winters | 0.903 | 0.932 | 0.965 | 7.665 | 3.672 | 1.363 | 4 |
| Auto ARIMA | 1.035 | 1.433 | 1.197 | 8.834 | 4.176 | 2.002 | 5 |
| Holt's Linear | 1.056 | 1.587 | 1.261 | 9.003 | 4.231 | 2.207 | 6 |
| Naïve | 1.152 | 1.873 | 1.369 | 9.822 | 4.593 | 2.392 | 7 |
| Theta | 1.254 | 1.936 | 1.391 | 10.615 | 5.679 | 2.367 | 8 |
| Drift | 1.243 | 3.101 | 1.761 | 10.636 | 3.672 | 3.238 | 9 |

As seen from Table 3, the univariate models performed mediocrely with average SMAPE scores hovering above 5. Furthermore, all the models have high Max Errors, showing clear areas of improvement. Figure 9 depicts the predictions of the models.

**Figure 9.** Predicted CO2 emissions

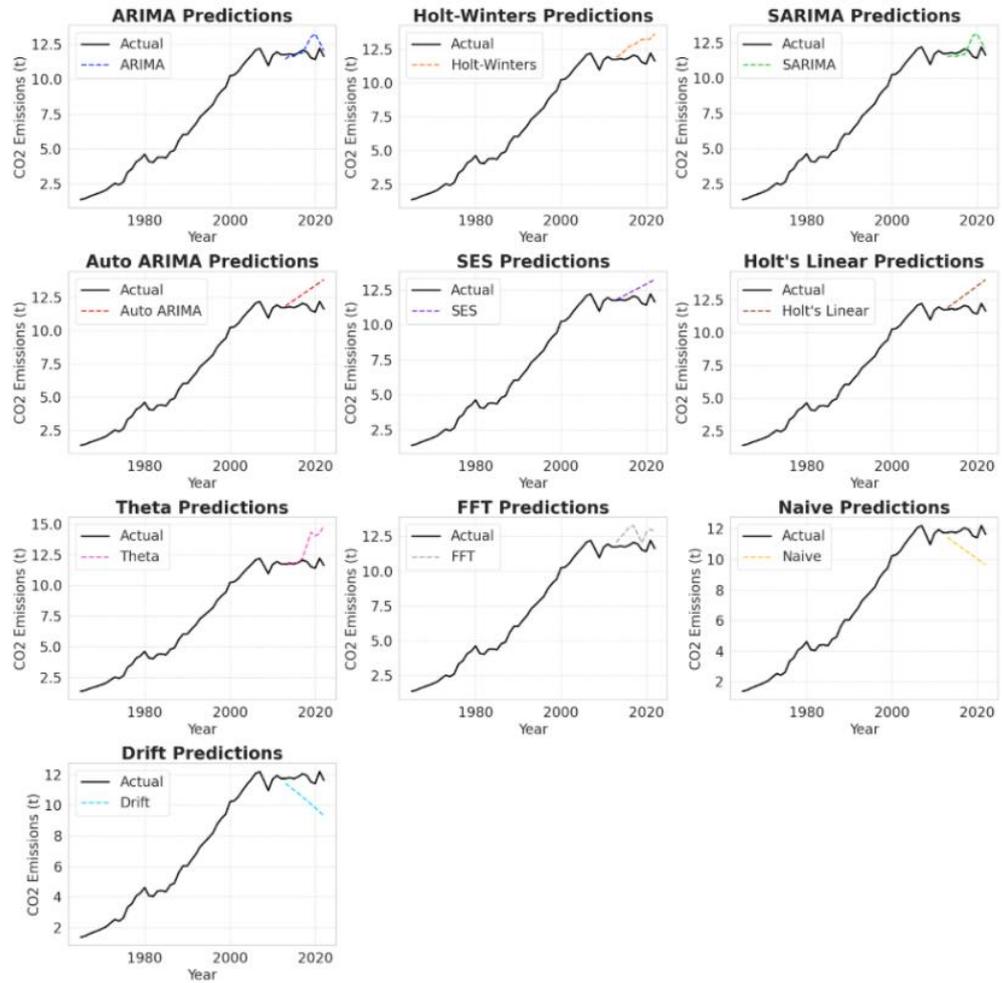

As seen in Figure 9, Most univariate time series models assumed linearity in Taiwan's CO2 trend. Without understanding other features and external factors, most models predicted a continuous growth in Taiwan's CO2 emissions. This indicates that the CO2 trend in Taiwan is more complex than expected. Significantly influenced by external factors like governmental policies, trade, and industrial activities, the CO2 trend in Taiwan requires more complex models to forecast.

*4.2. Multivariate Results*

Unlike univariate models, multivariate models are trained with data on annual Gas, Coal, and Oil consumption. Table 4 presents the performances of the multivariate models.

**Table 4.** Multivariate model performances.

| Models | MAE | MSE | RMSE | MAPE | SMAPE | Max Error | Rank |
|---|---|---|---|---|---|---|---|
| **FFNN** | **0.189** | **0.064** | **0.254** | **1.589** | **1.606** | **0.531** | **1** |
| **SVR** | **0.229** | **0.062** | **0.248** | **1.940** | **0.974** | **0.431** | **2** |
| **RFR** | **0.488** | **0.401** | **0.633** | **4.19** | **4.05** | **1.32** | **3** |
| Elastic Net Regression | 0.616 | 0.643 | 0.802 | 5.277 | 5.074 | 1.669 | 4 |
| LSTM | 0.927 | 1.210 | 1.131 | 7.903 | 7.498 | 1.899 | 5 |
| Decision Tree | 0.995 | 1.721 | 1.312 | 8.518 | 7.940 | 2.562 | 6 |
| BVAR | 1.110 | 2.083 | 1.443 | 9.422 | 8.719 | 2.868 | 7 |
| Ridge Regression | 1.109 | 2.021 | 1.421 | 9.491 | 8.811 | 2.734 | 8 |
| VECM | 1.130 | 2.085 | 1.444 | 9.660 | 8.961 | 2.767 | 9 |
| DFM | 1.131 | 2.088 | 1.445 | 9.668 | 8.967 | 2.769 | 10 |

The multivariate time series models achieved promising results, with significantly lower evaluation errors compared to univariate time series models. Specifically, the top multivariate model achieved an SMAPE of 1.606 compared to the 4.695 achieved by the top univariate model. Figure 10 depicts the predictions of the models.

**Figure 10.** Predicted CO2 emissions

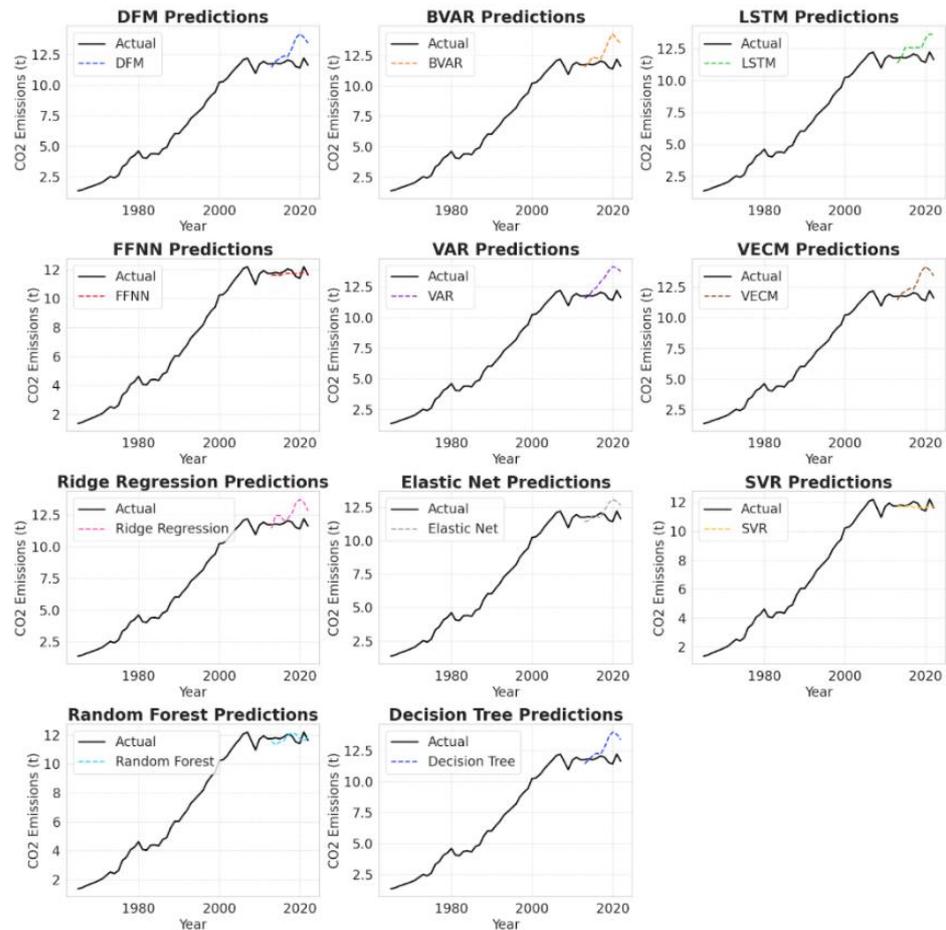

As seen from Figure 10, though some models, like the Decision Tree and BVAR, still predicted a continuous upward trend, more complex models like RFR and FFNN were able to accurately capture the feature-target relationships.

### 4.3. Ensemble Model Results

After comparing the results of the univariate and multivariate time series models, we designed an ensemble model to integrate the top-performers. Its design allows it to avoid extreme values and follow the general trend predicted by the base models. Linear Regression's role as the meta-model effectively assigns greater weights to more accurate models and less emphasis on models with poorer performances. Table 5 shows a comprehensive evaluation of the performance of the hybrid model.

**Table 5.** Ensemble model performance

| Models | MAE | MSE | RMSE | MAPE | SMAPE | Max Error |
|---|---|---|---|---|---|---|
| **Ensemble** | **0.166** | **0.049** | **0.222** | **1.398** | **1.407** | **0.531** |
| FFNN | 0.189 | 0.064 | 0.254 | 1.589 | 1.606 | 0.531 |
| SVR | 0.229 | 0.062 | 0.248 | 1.940 | 0.974 | 0.431 |
| Random Forest | 0.488 | 0.401 | 0.633 | 4.19 | 4.05 | 1.32 |

As seen from Table 5, the proposed ensemble model achieved higher accuracies compared to each of the top models. Figure 11 shows the graphical representation of the performance.

**Figure 11.** Predicted CO2 emissions

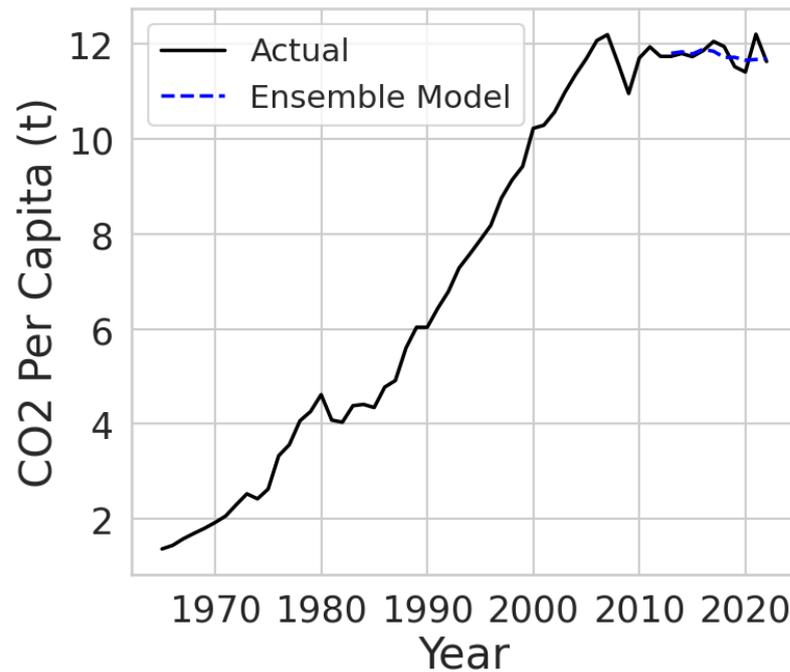

As seen from Figure 11, the proposed ensemble model is capable of predicting even the slightest fluctuations in emission trends. Therefore, this model was employed for a 10-year emission forecast.

*4.4. Future Forecasts*

After constructing the ensemble model, we shifted the dataset to forecast the CO2 emissions in Taiwan for the next 10 years. The model was trained on the entire dataset with annual samples from 1967 to 2022. Table 6 shows the numerical predictions from the model.

**Table 6.** CO2 forecasts

| Years | CO2 Per Capita (t) |
|---|---|
| 2023 | 11.475 |
| 2024 | 11.485 |
| 2025 | 11.874 |
| 2026 | 11.878 |
| 2027 | 11.980 |
| 2028 | 11.992 |
| 2029 | 11.821 |
| 2030 | 11.515 |
| 2031 | 11.745 |
| 2032 | 11.523 |

**Figure 12.** Projected CO2 emissions for the next 10 years.

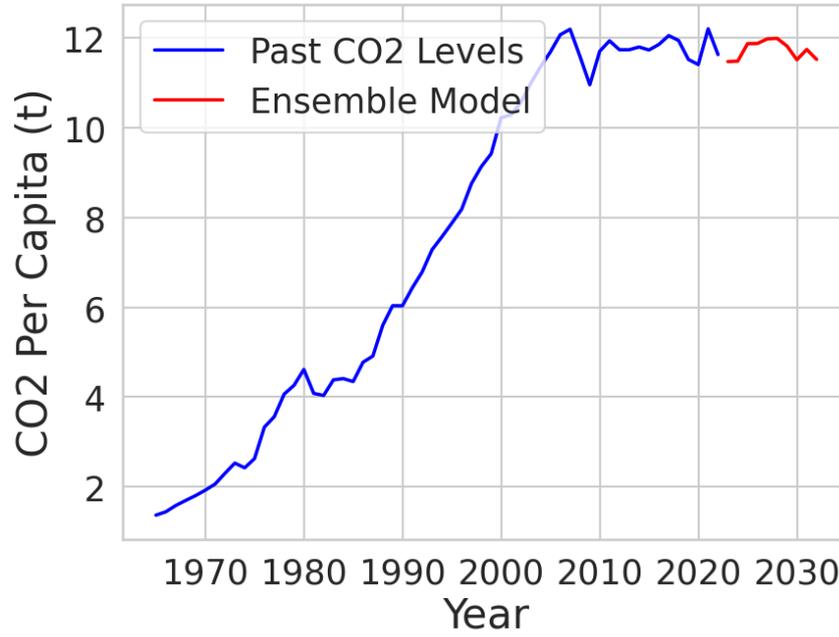

As seen in Figure 12, the forecast period includes the years 2023 and 2024 due to the lack of available data in our dataset. Regardless, emissions are projected to decline slightly toward the end of the forecast period. This shows promising results of the government's policies and measures in reducing CO2 emissions over time.

## 5. Conclusions

Successful forecasting of CO2 emissions is extremely beneficial for policymakers and business corporations to make informed decisions. This paper presents a comparative study of univariate and multivariate models; multivariate time series models have been shown to produce higher accuracies due to the interconnectedness between the features and the target. To further enhance forecasting accuracy, a novel ensemble model is proposed. The model combines the predictions from FFNN, SVR, and RFR using Linear Regression as the meta-model. It utilizes the variations in the predictions to ensure robust and accurate forecasts. This hybrid model is superior compared to the 10 univariate and 11 multivariate time series models presented in this paper. Furthermore, it is also shown to be more accurate than many previously proposed models [7, 8, 9, 10, 11, 12, 13]. The accuracy of our model can be attributed to the thoroughness of our data preprocessing and our ensemble model design. Moreover, this paper presents the future forecast of Taiwan's CO2 emissions for the next 10 years. The results of this paper demonstrate the power of machine learning in forecasting environmental trends, showing great potential for future work. Future research can employ more complex time series models such as Prophet, Bayesian Structural Time Series (BSTS), and Gaussian Processes (GP), and compare the results with the models presented in this study. Furthermore, more datasets can be used to verify the performance of the proposed ensemble model under different climate conditions.